\documentclass[letterpaper]{article} 
\usepackage{aaai25}  
\usepackage{times}  
\usepackage{helvet}  
\usepackage{courier}  
\usepackage[hyphens]{url}  
\usepackage{graphicx} 
\usepackage{natbib}  
\usepackage{caption}
\usepackage{amssymb}
\usepackage{enumitem}
\usepackage{array}
\usepackage{pgfplots}
\pgfplotsset{compat=1.18}
\usepackage{multirow}
\usepackage{array}
\usepackage{booktabs}
\usepackage{amsmath} 
\usepackage{algorithm}
\usepackage{algorithmic}

\usepackage{newfloat}
\usepackage{listings}
\DeclareCaptionStyle{ruled}{labelfont=normalfont,labelsep=colon,strut=off} 
\floatstyle{ruled}
\newfloat{listing}{tb}{lst}{}
\floatname{listing}{Listing}
\usepackage{colortbl}
\usepackage{xcolor}

\title{LoRA-Mini : Adaptation Matrices Decomposition and Selective Training}
\author{
   Ayush Singh,Rajdeep Aher,Shivank Garg }

\affiliations{
    Vision and Language Group, Indian Institute of Technology Roorkee, Roorkee, Uttarakhand, India - 247667 \\
   ayush\_s@mt.iitr.ac.in ,  aher\_rp@ma.iitr.ac.in , shivank\_g@mfs.iitr.ac.in   
}

\usepackage{bibentry}

\begin{document}

\maketitle

\begin{abstract}
The rapid advancements in large language models (LLMs) have revolutionized natural language processing, creating an increased need for efficient, task-specific fine-tuning methods. Traditional fine-tuning of LLMs involves updating a large number of parameters, which is computationally expensive and memory-intensive. Low-Rank Adaptation (LoRA) has emerged as a promising solution, enabling parameter-efficient fine-tuning by reducing the number of trainable parameters. However, while LoRA reduces the number of trainable parameters, LoRA modules still create significant storage challenges. We propose LoRA-Mini, an optimized adaptation of LoRA that improves parameter efficiency by splitting low-rank matrices into four parts, with only the two inner matrices being trainable. This approach achieves upto a 20x reduction compared to standard LoRA in the number of trainable parameters while preserving performance levels comparable to standard LoRA, addressing both computational and storage efficiency in LLM fine-tuning.

\end{abstract}

%


\section{Introduction}

The rapid growth of large language models (LLMs) such as GPT-3 \cite{brown2020languagemodelsfewshotlearners}, GPT-4 \cite{openai2024gpt4technicalreport}, Llama \cite{dubey2024llama3herdmodels}, and Mistral \cite{jiang2023mistral7b} has led to various advancements in the field of natural language processing, enabling LLMs to achieve high performance across various benchmarks such as IFEval \cite{zhou2023instructionfollowingevaluationlargelanguage}, SQuAD \cite{rajpurkar-etal-2016-squad} etc. However, the computational and memory costs involved in the training of these models from scratch are substantial, posing significant challenges. Fine-tuning \cite{zhang2024instructiontuninglargelanguage} is a popular method to adapt pre-trained models for specific downstream tasks, but requires extensive resources when tuning all the model parameters, making it impractical for many applications. To address these limitations, researchers developed Parameter-Efficient Fine-Tuning (PEFT) \cite{han2024parameterefficientfinetuninglargemodels} techniques, which limit the number of adjustable parameters during fine-tuning, achieving accuracy comparable to full finetuning while significantly reducing both computational and memory costs. 
Among PEFT approaches, Low-Rank Adaptation(LoRA)  \cite{hu2021loralowrankadaptationlarge} has emerged as a widely adopted technique. The technique adapts large models by adding trainable low-rank matrices to the model’s layers. It builds on the observation that fine-tuning updates often have low intrinsic dimensionality. By incorporating this, LoRA achieves high efficiency without a major compromise in model performance, allowing for more scalable adaptation of LLMs to diverse tasks. These advances highlight LoRA’s role in balancing efficiency and effectiveness, enabling widespread application of large-scale models in resource-constrained environments. 


Our approach builds on existing Parameter-Efficient Fine-Tuning (PEFT) methods by targeting an even lower parameter count, aiming to maintain model performance while further minimizing memory usage.

Although deep networks may have a vast number of parameters, but only a small number of parameters significantly impact the learning process. Hence, we refine the intrinsic rank within the current LoRA framework by decomposing the standard matrices \( A \) and \( B \) into two sub-matrices each, with only one matrix from each pair being trainable. 

Our experiments cover a range of models, including BERT \cite{devlin2019bertpretrainingdeepbidirectional}, RoBERTa \cite{liu2019robertarobustlyoptimizedbert}, and T5 \cite{raffel2023exploringlimitstransferlearning}, with a primary focus on performance metrics from the GLUE \cite{wang2019gluemultitaskbenchmarkanalysis} benchmark and English-Romanian translation task from WMT16 \cite{bojar-EtAl:2016:WMT1} dataset. Our results show that LoRA-Mini attains accuracy comparable to full fine-tuning approaches while significantly reducing memory requirements. Our key contributions include:
\begin{enumerate} 
    \item Evaluating the effectiveness of selectively freezing parameters within LoRA layers while maintaining model quality.
    \item Minimizing the number of trainable parameters relative to earlier PEFT methods, while achieving performance on par with full fine-tuning and LoRA.
    \item Evaluating the scalability of our approach on diverse set of tasks and models.
\end{enumerate}

\begin{figure*}[ht]
\centering
\includegraphics[scale=0.20]{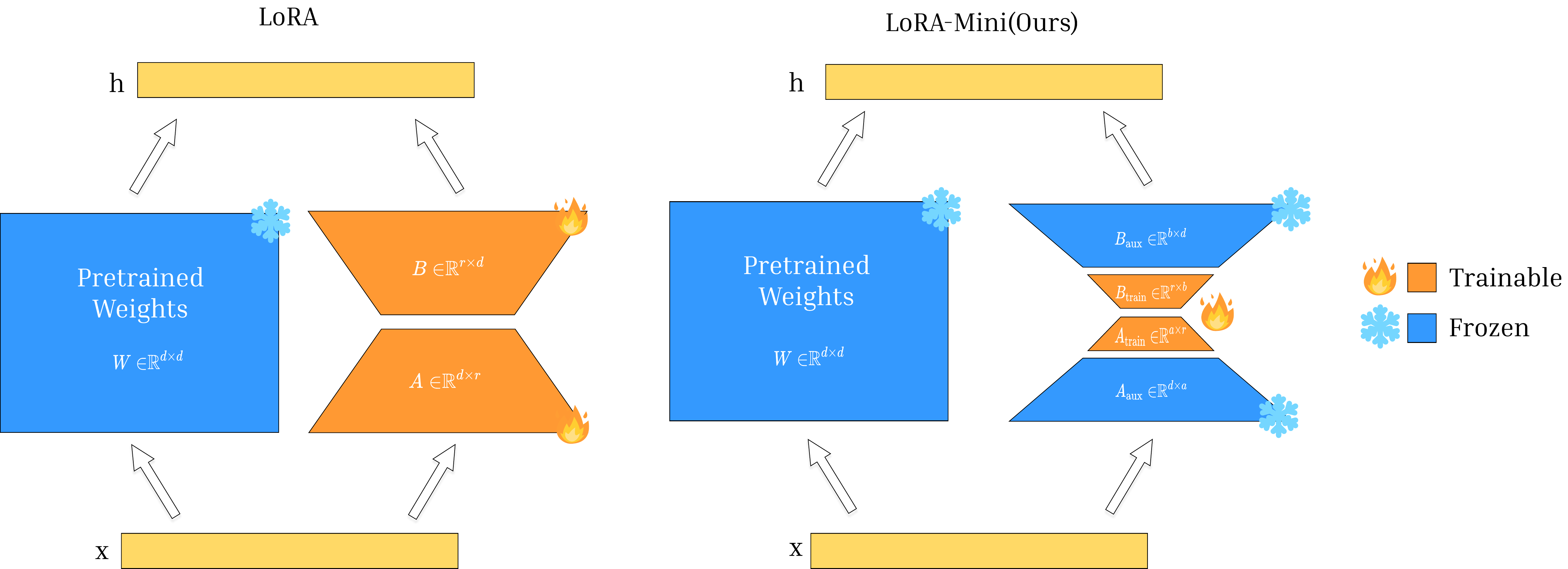}
\caption{A visual comparison of the LoRA and LoRA-Mini techniques.}
\end{figure*}

\section{Related Work}
Parameter-efficient fine-tuning has emerged as an essential approach for adapting large language models (LLMs) without incurring high computational and memory costs. Techniques such as Adapter-based methods \cite{houlsby2019}, Prefix-tuning \cite{li2021}, and Prompt-tuning \cite{lester2021} provide alternatives to conventional fine-tuning, allowing for task-specific adaptation by inserting trainable parameters at selective layers. Adapter-based methods introduce small bottleneck layers to facilitate training while keeping core model parameters fixed. However, most adapter layers can add inference latency, as they must be processed sequentially due to their additional depth within each Transformer block. Variants of adapters attempt to mitigate this, such as by using a single adapter layer per block with an additional LayerNorm \cite{lin2020}, or by reducing latency through layer pruning and multi-task adaptation \cite{ruckle2020, pfeiffer2021}.

On the other hand, prompt-based methods like Prefix-tuning face optimization challenges. Optimizing prompt-based adaptations can be difficult, with performance sensitivity to the number of trainable parameters and prone to non-monotonic behavior \cite{li2021}. Additionally, reserving part of the sequence length for adaptation may reduce the effective input length available for the downstream task, potentially limiting performance. 

Consequently, LoRA \cite{hu2021} provides an efficient approach to fine-tuning large language models by focusing on low-rank updates to the model weights rather than modifying the model parameters directly or adjusting input sequences. Unlike adapter-based methods, which add additional layers that can introduce latency in inference, or prompt-based methods, which often struggle with prompt optimization, it preserves the model’s original architecture and sequence length, ensuring that the models' structure remains unchanged during deployment. Nevertheless, it faces a significant drawback: the large memory overhead associated with storing low-rank matrices, which presents challenges for deployment in resource-constrained environments.

Recent works have tried improving the original LoRA framework for efficient language model adaptation. Some significant works include LoRA-FA \cite{zhang2023lorafamemoryefficientlowrankadaptation} which achieves 1.4x memory reduction by freezing the projection-down matrix and training only the projection-up matrix. Bayesian-LoRA \cite{yang2024bayesianlowrankadaptationlarge} further enhances efficiency by dynamically allocating ranks to layers based on data requirements.

Our technique, however, introduces additional parameter reduction from previous methods by decomposing matrices $A$ and $B$ into auxiliary and trainable components $(A_{aux}, A_{train}, B_{train}, B_{aux})$, with only the middle matrices being trainable. This substantially reduces the parameter count.
\definecolor{gold}{rgb}{0.84, 1, 0.84} 
\definecolor{silver}{rgb}{0.84, 0.93, 1} 

\definecolor{bronze}{rgb}{1, 1, 1}

\begin{table*}[h!]
    \centering
    \setlength{\tabcolsep}{10pt} 
    \begin{tabular}{lcccccc}
    \toprule
    \textbf{Method} & \textbf{Parameters} & \textbf{Rank} & \textbf{STSB} & \textbf{COLA} & \textbf{MRPC} & \textbf{RTE}\\
\midrule
    \textbf{}  FFT & 125M & - & \cellcolor{gold}\textbf{90.77} & 80.54 & \cellcolor{gold}\textbf{89.46} & \cellcolor{silver}\textit{74.01}\\
    \textbf{}  LoRA & 0.90M & 8 & 88.49 & 80.53 & \cellcolor{bronze} 87.50 & 62.81\\
    \textbf{}  LoRA & 1.80M & 16 & 89.11 & 80.24 & \cellcolor{bronze} 87.50 & 67.14\\
    \textbf{}  LoRA & 3.50M & 32 & \cellcolor{bronze} \cellcolor{bronze} 89.87 & \cellcolor{bronze}82.35 & 85.78 & 68.95\\
\midrule

    \textbf{}  Ours(D)& 0.04M & 8 & 89.62 & 81.00 & 86.52 & 69.31\\
    \textbf{}  Ours(D)& 0.07M & 16 & 89.51 & 81.50 & \cellcolor{bronze}\cellcolor{silver}\textit{ 88.23} & \cellcolor{bronze}72.92\\
    \textbf{}  Ours(D)& 0.15M & 32 & 89.55 & 82.17 & 86.76 & \cellcolor{gold}74.72\\
    \textbf{}  Ours(D+A)& 0.08M & 8 & 89.62 & \cellcolor{bronze}82.35 & 87.01 & 70.75\\
    \textbf{}  Ours(D+A)& 0.15M & 16 & 89.67 & \cellcolor{silver}\textit{82.84} & 86.02 & 69.31\\
    \textbf{}  Ours(D+A)& 0.30M & 32 & \cellcolor{silver}\textit{90.36} & \cellcolor{gold}\textbf{83.00} & 86.52 & 68.23\\
    \bottomrule
    \end{tabular}
    \caption{In this experiment, we use $\textbf{RoBERTa}_{\text{base}}$. We report Pearson correlation for the STSB task and accuracy for the remaining tasks. For all metrics, higher values indicate better performance. Green(or bold) entries denote the best and Blue(or italics) denote the second best entries of a column}
    \label{tab:roberta}
\end{table*}

\begin{table*}[h!]
    \centering
    \label{tab:bert}
    \setlength{\tabcolsep}{10pt} 
    \begin{tabular}{lcccccc}
    \toprule
    \textbf{Method} & \textbf{Parameters} & \textbf{Rank} & \textbf{STSB} & \textbf{COLA} & \textbf{MRPC} & \textbf{RTE}\\
\midrule
    \textbf{}  FFT & 110M & - & \cellcolor{gold} \textbf{88.74} & \cellcolor{silver}\textit{ 81.69} &  82.11 & 64.98 \\
    \textbf{}  LoRA & 0.90M & 8 & 86.44 & \cellcolor{silver} \textit{81.69} & 76.47 & 63.90 \\
    \textbf{}  LoRA & 1.80M & 16 & 87.00 & 79.96 & 81.62 & 61.73 \\
    \textbf{}  LoRA & 3.50M & 32 & 86.96 & 79.29 & 80.15 & \cellcolor{gold} \textbf{68.95 }\\
\midrule
    \textbf{}  Ours(D)& 0.04M & 8 & 86.21 & \textbf{\cellcolor{gold} 81.78} & \cellcolor{bronze} 82.33& \cellcolor{silver}\textit{ 67.87}\\

    \textbf{}  Ours(D)& 0.07M & 16 & \cellcolor{silver}\textit{ 88.35} & 80.82& \cellcolor{silver}\textit{ 83.57}& 64.26\\
    \textbf{}  Ours(D)& 0.15M & 32 & 87.13 & \cellcolor{bronze} 81.40 &  82.59& 61.01\\
    \textbf{}  Ours(D+A)& 0.08M & 8 & 88.02 & 79.29 & 78.92& \cellcolor{bronze} 67.14\\
    \textbf{}  Ours(D+A)& 0.15M & 16 & \cellcolor{bronze} 88.25 & 79.58 & 81.37 &64.26\\
    \textbf{}  Ours(D+A)& 0.30M & 32  & 87.98 & 80.82 & \textbf{\cellcolor{gold} 84.31} & 65.70\\

    \bottomrule
    \end{tabular}
    \caption{In this experiment, we use $\textbf{BERT}_{\text{base}}$ : We report Pearson correlation for the STSB task and accuracy for the remaining tasks. For all metrics, higher values indicate better performance. Green(or bold) entries denote the best and Blue(or italics) denote the second best entries of a column}
    \label{tab:bert}
\end{table*}
\section{Methodology}


LoRA-Mini focuses on the introduction of selective training within decomposed matrices. We divide each LoRA matrix into a trainable and frozen part, allowing us to limit the update space and execute controlled updates, where the frozen outer matrices guide the training process.
The mathematical formulation of our approach can be expressed as follows:
\newline
Let $AB \in \mathbb{R}^{d \times k}$ represent our target weight matrix where d is input and k is output dimension. We decompose the LoRA matrices into:

\begin{itemize}
    \item Outer auxiliary matrices: \( A_{\text{aux}} \in \mathbb{R}^{d \times \text{a}} \), \( B_{\text{aux}} \in \mathbb{R}^{\text{b} \times k} \) (frozen)
    \item Inner trainable matrices: \( A_{\text{train}} \in \mathbb{R}^{a \times r} \), \( B_{\text{train}} \in \mathbb{R}^{r \times b} \) (trainable)
\end{itemize}

Let x be the input to a layer. Then the output h from the layer after applying LoRA-Mini becomes : 
\begin{equation}
h = \left(W + A_{\text{aux}} \cdot A_{\text{train}} \cdot B_{\text{train}} \cdot B_{\text{aux}}\right) \cdot x
\end{equation}

In LoRA, given a pre-trained weight matrix \( W \in \mathbb{R}^{d \times k} \), the weight update \( \Delta W \) is through two matrices, \( A \in \mathbb{R}^{d \times r} \) and \( B \in \mathbb{R}^{r \times k} \), where \( r \ll \min(d, k) \). This parameterization leads to \( \Delta W = AB \), resulting in \( r \times (d + k) \) trainable parameters, reducing the training overhead.

Our method extends this reduction further by limiting the trainable parameters to \( r \times (a + b) \), where \( a \) and \( b \) are predefined dimensional constraints, as detailed in later sections. 

All matrices are initialized using the Kaiming method, with a Kaiming coefficient of \( \sqrt{5} \), ensuring weight variance is scaled appropriately for stable gradient propagation. 
Our approach yields several advantages. First, by constraining updates to the inner matrices, while keeping outer matrices fixed, we achieve substantial memory savings without sacrificing the models efficiency. Second, the weight update, ($\Delta W = A_{aux}A_{train}B_{train}B_{aux}$) operates within a controlled subspace(within inner matrices), enabling efficient parameter optimization while maintaining model stability.

\section{Experiments and Results}
To do a robust evaluation of our approach, we broadly divided our experiment into two categories. 

First, we evaluate our approach on the NLU and NLI tasks using the GLUE Benchmark \cite{wang2019gluemultitaskbenchmarkanalysis} on RoBERTa \cite{liu2019robertarobustlyoptimizedbert} and BERT \cite{devlin2019bertpretrainingdeepbidirectional}. We particularly chose four tasks from the GLUE Benchmark: STSB \cite{stsb}, CoLA \cite{cola}, MRPC \cite{mrpc}, and RTE \cite{rte}. The experiments were conducted using the following configurations : 
\begin{itemize}
    \item Full fine-tuning of the entire model
    \item Standard LoRA applied to dense and attention layers (rank values $r=4$, $8$, $16$)
    \item LoRA-Mini applied only to dense layers
    \item LoRA-Mini applied to both dense and attention layers
\end{itemize}

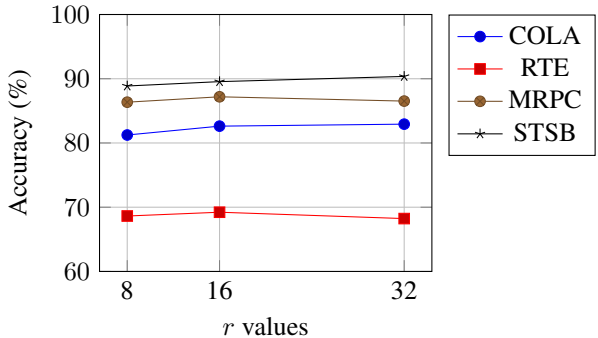
\begin{figure}[H]  
    \centering
    \begin{tikzpicture}
    \begin{axis}[
        width=6cm,   
        height=5cm,   
        xlabel={$r$ values},
        ylabel={Accuracy (\%)},
        xtick={8, 16, 32},
        xticklabels={8, 16, 32},
        legend style={at={(1.05,1)}, anchor=north west},
        ymin=60, ymax=100,
        grid=both,
    ]

    \addplot coordinates {(8, 81.24) (16, 82.62) (32, 82.93)};
    \addlegendentry{COLA}

    \addplot coordinates {(8, 68.63) (16, 69.22) (32, 68.23)};
    \addlegendentry{RTE}

    \addplot coordinates {(8, 86.35) (16, 87.19) (32, 86.52)};
    \addlegendentry{MRPC}

    \addplot coordinates {(8, 88.87) (16, 89.55) (32, 90.36)};
    \addlegendentry{STSB}

    \end{axis}
    \end{tikzpicture}
    
    \caption{Accuracy comparison across different values of $r$ for four tasks using LoRA-Mini applied to the Attention and Dense layers of RoBERTa-Base. The graph reports average of all combinations of a and b in that rank.}
    \label{fig:accuracy_vs_r}  
\end{figure}

We tested rank values of 8, 16, and 32, with $a$ and $b$ as 8, 16, 32, and 64. This enabled an effective comparison of LoRA-Mini's impact on feed-forward versus attention mechanisms. We did not experiment with applying our approach exclusively to attention layers, as previous research  \cite{geva2021transformerfeedforwardlayerskeyvalue} has demonstrated that this would not yield significant improvements in overall accuracy. From our experiments, we observed that the configuration with $a$ and $b$ as 64 performs consistently well. Thus, we reported results with varying ranks and keeping $a$ and $b$ constant in the main paper. From Table~\ref{tab:roberta} and Table~\ref{tab:bert}, we can see that our approach performs consistently at par or greater than full-fine-tuning and LoRA. A comparison of rank with accuracy when LoRA-Mini is applied to both dense and attention layers of RoBERTa is shown in Figure~\ref{fig:accuracy_vs_r}.

We also evaluate the performance of our approach on generative tasks, with a particular focus on language translation. It is a significant benchmark for assessing model performance in real-world applications, as it involves both understanding and generating human language. To better understand how our approach scales with increasing model size, we select two models with different capacities: T5-Small and T5-Base. We fine-tuned these models on the English-Romanian subset of the WMT16 dataset, a widely used resource for machine translation tasks.

 We then evaluated the generated translations using the BLEU \cite{papineni-etal-2002-bleu} and ROUGE-L \cite{lin-2004-rouge} metrics. The experimental setup closely follows the previous set, but in this case, we applied LoRA-Mini to both the attention and dense layers of the models. The detailed results from these evaluations are presented in Table~\ref{T5-small} and Table~\ref{T5-base}. These show that LoRA-Mini consistently matches the performance of both base LoRA and full fine-tuning on both BLEU and ROUGE-L metrics. 

These experiments clearly indicate that LoRA-Mini is a reliable approach for both language generation and understanding tasks. It not only delivers competitive results in these tasks but also reduces the parameter count considerably compared to previous methods, which can lead to improved efficiency and scalability. Results of all configurations and additional analysis are provided in the Appendix.



\begin{table}[H]
    \centering
    \setlength{\tabcolsep}{4pt} 
    \begin{tabular}{lccccc}
    \toprule
    \textbf{Method} & \textbf{Params} & \textbf{Rank} & \textbf{BLEU} & \textbf{ROUGEL}\\
    \midrule
    \textbf{}  FFT & 60M & - &26.11 & 45.95 &\\
    \textbf{}  LoRA & 0.14M & 1 & 26.17 & 45.86\\
    \textbf{}  LoRA & 0.29M & 2 & 26.1 & 46.04\\
    \textbf{}  LoRA & 0.59M & 4 & 26.1 & 46.05\\
    \textbf{}  LoRA & 1.2M & 8 & 26.11 & 45.87\\
    \textbf{}  LoRA & 2.3M & 16 & 26.08 & 45.91\\
\midrule
    \textbf{}  Ours & 0.08M & 8 & 26.01 & 45.62\\
    \textbf{}  Ours & 0.16M & 16 & 26.02 & 45.88\\
    \textbf{}  Ours & 0.32M & 32 & 25.82 & 45.54\\
    \bottomrule
    \end{tabular}
    \caption{In this experiment, we use \textbf{T5 Small}. We report BLEU, and ROUGEL. For all metrics, higher values indicate better performance.}
    \label{T5-small}
\end{table}

\begin{table}[H]
    \centering
    \setlength{\tabcolsep}{4pt} 
    \begin{tabular}{lccccc}
    \toprule
    \textbf{Method} & \textbf{Params} & \textbf{Rank} & \textbf{BLEU} & \textbf{ROUGEL}\\
    \midrule
    \textbf{}  FFT & 223M & - & 27.27 & 47.20 &\\
    \textbf{}  LoRA & 0.38M & 1 & 26.89 & 47.08\\
    \textbf{}  LoRA & 0.76M & 2 & 27.12 & 47.23\\
    \textbf{}  LoRA & 1.50M & 4 & 27.01 & 47.21\\
    \textbf{}  LoRA & 3.0M & 8 & 27.02 & 47.14\\
    \textbf{}  LoRA & 6.1M & 16 & 26.90 & 47.08\\
\midrule
    \textbf{}  Ours & 0.16M & 8 & 26.83 & 46.98\\
    \textbf{}  Ours & 0.32M & 16 & 26.99 & 46.77\\
    \textbf{}  Ours & 0.64M & 32 & 26.92 & 46.78\\
    \bottomrule
    \end{tabular}
    \caption{In this experiment, we use \textbf{T5 Base}. We report BLEU and ROUGEL. For all metrics, higher values indicate better performance.}
    \label{T5-base}
\end{table}
\section{Conclusion}
We present LoRA-Mini as a parameter-efficient fine-tuning approach, demonstrating significant improvements in memory efficiency and task performance. Our results indicate that LoRA-Mini can achieve comparable or superior results compared to LoRA and full fine-tuning while substantially reducing the number of trainable parameters across a wide range of tasks, making it an ideal solution for finetuning LLMs in resource-constrained environments. 


\section{Limitations and Future Works}
Due to computational constraints, we were unable to evaluate our LoRA approach on larger tasks of GLUE benchmark as well as on larger models such as LLaMA3.1-8B \cite{dubey2024llama3herdmodels}, Mistral-7B \cite{jiang2023mistral7b} etc, or on more extensive benchmarks datasets like MMLU \cite{hendryckstest2021}, Math10k \cite{hu2023llm}, COMMONSENSE170K \cite{hu2023llm} etc. Expanding to these larger models and benchmarks could provide a more comprehensive understanding of our method’s scalability and performance in diverse domains and model sizes.

Future research could explore alternative matrix decomposition methods, such as partitioning LoRA matrices into more smaller components and experimenting with various combinations of trainable and frozen sections. Another approach might involve freezing randomly selected matrices \cite{zhu2024asymmetrylowrankadaptersfoundation}. Techniques like QR decomposition or singular value decomposition (SVD) could be used to initialize these matrices, potentially enhancing performance. Furthermore, this approach could be tested to approximate feedforward networks (FFNs) \cite{zeng2024expressivepowerlowrankadaptation}. Investigating a latent representation between LoRA matrices might also increase parameter efficiency.

\bibliography{aaai25}

\clearpage

\appendix
\subsection{Experimental Setup Details}
\textbf{Models : } 

\begin{itemize}[left=0pt]
    \item \textbf{BERT} \footnote{huggingface.co/google-bert/bert-base-uncased}- A bidirectional transformer model with 110 million parameters, pre-trained on masked language modeling and designed for natural language understanding tasks.

    \item \textbf{RoBERTa Base} \footnote{huggingface.co/FacebookAI/roberta-base}- An optimized version of BERT with 125 million parameters, featuring more extensive pre-training, improving performance on a wide range of language tasks.

    \item \textbf{T5-Small} \footnote{huggingface.co/google-t5/t5-small}- A smaller version of the Text-To-Text Transfer Transformer (T5) with 60 million parameters, optimized for efficiency on generative tasks.

    \item \textbf{T5-Base} \footnote{huggingface.co/google-t5/t5-base}- The base variant of T5 with 223 million parameters.
\end{itemize}

\textbf{Benchmarks and Datasets : }
\begin{itemize}[left=0pt]
    \item \textbf{GLUE Dataset} \footnote{huggingface.co/datasets/nyu-mll/glue}:
    GLUE, or General Language Understanding Evaluation, is a benchmark designed to assess language understanding models across diverse NLP tasks, including CoLA (Corpus of Linguistic Acceptability), SST-2 (Stanford Sentiment Treebank), MRPC (Microsoft Research Paraphrase Corpus), STS-B (Semantic Textual Similarity Benchmark), QQP (Quora Question Pairs), MNLI (Multi-Genre Natural Language Inference), QNLI (Question Natural Language Inference), RTE (Recognizing Textual Entailment), and WNLI (Winograd NLI). Of these, we only train on 4 tasks, namely STSB, COLA, MRPC, and RTE.

    \item \textbf{WMT Dataset} \footnote{huggingface.co/datasets/wmt/wmt16}:
    The WMT-16 (Workshop on Machine Translation 2016) benchmark is a standardized evaluation framework for comparing translation models across multiple language pairs using reliable metrics like BLEU. In this research, we use the Romanian-to-English subset of WMT-16 to evaluate our model’s translation accuracy and generalization across linguistic structures.

\end{itemize}


\clearpage
\begin{table*}
\centering
\begin{tabular}{|c|c|c|c|c|c|c|c|c|}
\hline
\textbf{r} & \textbf{a} & \textbf{b} & \textbf{Accuracy(D)} & \textbf{Parameters(D)} & \textbf{Percentage(D)} & \textbf{Accuracy(D+A)} & \textbf{Parameters(A+D)} & \textbf{Percentage(A+D)} \\
\hline
8 & 16 & 16 & 77.661\% & 11010 & 0.009\% & 80.058\% & 20226 & 0.016\% \\
8 & 16 & 32 & 79.866\% & 15746 & 0.013\% & 81.016\% & 29570 & 0.024\% \\
8 & 16 & 64 & 81.687\% & 25218 & 0.020\% & 81.592\% & 48258 & 0.039\% \\
8 & 32 & 16 & 80.058\% & 15746 & 0.013\% & 80.345\% & 29570 & 0.024\% \\
8 & 32 & 32 & 80.729\% & 20482 & 0.016\% & 80.345\% & 38914 & 0.031\% \\
8 & 32 & 64 & 80.633\% & 29954 & 0.024\% & 82.934\% & 57602 & 0.046\% \\
8 & 64 & 16 & 78.428\% & 25218 & 0.020\% & 79.866\% & 48258 & 0.039\% \\
8 & 64 & 32 & 81.016\% & 29954 & 0.024\% & 82.646\% & 57602 & 0.046\% \\
8 & 64 & 64 & 81.016\% & 39426 & 0.032\% & 82.359\% & 76290 & 0.061\% \\
16 & 32 & 32 & 81.496\% & 39426 & 0.032\% & 82.359\% & 76290 & 0.061\% \\
16 & 32 & 64 & 82.071\% & 58370 & 0.047\% & 83.317\% & 113666 & 0.091\% \\
16 & 64 & 32 & 80.825\% & 58370 & 0.047\% & 81.975\% & 113666 & 0.091\% \\
16 & 64 & 64 & 81.496\% & 77314 & 0.062\% & 82.838\% & 151042 & 0.121\% \\
32 & 64 & 64 & 82.167\% & 153090 & 0.122\% & 82.934\% & 300546 & 0.240\% \\
\hline
\end{tabular}
\caption{RoBERTa-CoLA}
\end{table*}

\begin{table*}
\centering
\begin{tabular}{|c|c|c|c|c|c|c|c|c|}
\hline
\textbf{r} & \textbf{a} & \textbf{b} & \textbf{Accuracy(D)} & \textbf{Parameters(D)} & \textbf{Percentage(D)} & \textbf{Accuracy(A+D)} & \textbf{Parameters(A+D)} & \textbf{Percentage(A+D)} \\
\hline
8 & 16 & 16 & 67.148\% & 11010 & 0.009\% & 71.480\% & 20226 & 0.016\% \\
8 & 16 & 32 & 67.870\% & 15746 & 0.013\% & 70.036\% & 29570 & 0.024\% \\
8 & 16 & 64 & 70.036\% & 25218 & 0.020\% & 70.036\% & 48258 & 0.039\% \\
8 & 32 & 16 & 64.982\% & 15746 & 0.013\% & 68.592\% & 29570 & 0.024\% \\
8 & 32 & 32 & 67.148\% & 20482 & 0.016\% & 68.592\% & 38914 & 0.031\% \\
8 & 32 & 64 & 75.090\% & 29954 & 0.024\% & 64.982\% & 57602 & 0.046\% \\
8 & 64 & 16 & 65.704\% & 25218 & 0.020\% & 63.899\% & 48258 & 0.039\% \\
8 & 64 & 32 & 67.870\% & 29954 & 0.024\% & 69.314\% & 57602 & 0.046\% \\
8 & 64 & 64 & 69.314\% & 39426 & 0.032\% & 70.758\% & 76290 & 0.061\% \\
16 & 32 & 32 & 68.953\% & 39426 & 0.032\% & 68.592\% & 76290 & 0.061\% \\
16 & 32 & 64 & 71.119\% & 58370 & 0.047\% & 71.480\% & 113666 & 0.091\% \\
16 & 64 & 32 & 71.119\% & 58370 & 0.047\% & 67.509\% & 113666 & 0.091\% \\
16 & 64 & 64 & 72.924\% & 77314 & 0.062\% & 69.314\% & 151042 & 0.121\% \\
32 & 64 & 64 & 74.729\% & 153090 & 0.122\% & 68.231\% & 300546 & 0.240\% \\
\hline
\end{tabular}
\caption{RoBERTa-RTE}
\end{table*}

\begin{table*}
\centering
\begin{tabular}{|c|c|c|c|c|c|c|c|c|}
\hline
\textbf{r} & \textbf{a} & \textbf{b} & \textbf{Accuracy(D)} & \textbf{Parameters(D)} & \textbf{Percentage(D)} & \textbf{Accuracy(D+A)} & \textbf{Parameters(A+D)} & \textbf{Percentage(A+D)} \\
\hline
8 & 16 & 16 & 84.069\% & 11010 & 0.009\% & 86.029\% & 20226 & 0.016\% \\
8 & 16 & 32 & 84.804\% & 15746 & 0.013\% & 85.784\% & 29570 & 0.024\% \\
8 & 16 & 64 & 87.255\% & 25218 & 0.020\% & 85.539\% & 48258 & 0.039\% \\
8 & 32 & 16 & 83.578\% & 15746 & 0.013\% & 85.294\% & 29570 & 0.024\% \\
8 & 32 & 32 & 86.029\% & 20482 & 0.016\% & 87.500\% & 38914 & 0.031\% \\
8 & 32 & 64 & 87.500\% & 29954 & 0.024\% & 86.765\% & 57602 & 0.046\% \\
8 & 64 & 16 & 84.804\% & 25218 & 0.020\% & 86.275\% & 48258 & 0.039\% \\
8 & 64 & 32 & 85.539\% & 29954 & 0.024\% & 87.010\% & 57602 & 0.046\% \\
8 & 64 & 64 & 86.520\% & 39426 & 0.032\% & 87.010\% & 76290 & 0.061\% \\
16 & 32 & 32 & 87.745\% & 39426 & 0.032\% & 87.255\% & 76290 & 0.061\% \\
16 & 32 & 64 & 85.294\% & 58370 & 0.047\% & 88.480\% & 113666 & 0.091\% \\
16 & 64 & 32 & 85.784\% & 58370 & 0.047\% & 87.010\% & 113666 & 0.091\% \\
16 & 64 & 64 & 88.235\% & 77314 & 0.062\% & 86.029\% & 151042 & 0.121\% \\
32 & 64 & 64 & 86.765\% & 153090 & 0.122\% & 86.520\% & 300546 & 0.240\% \\
\hline
\end{tabular}
\caption{RoBERTa-MRPC}
\end{table*}

\begin{table*}
\centering
\begin{tabular}{|c|c|c|c|c|c|c|c|c|}
\hline
\textbf{r} & \textbf{a} & \textbf{b} & \textbf{Pearson (D)} & \textbf{Parameters (D)} & \textbf{Percentage (D)} & \textbf{Pearson (A+D)} & \textbf{Parameters (A+D)} & \textbf{Percentage (A+D)} \\
\hline
8 & 16 & 16 & 86.598\% & 10241 & 0.008\% & 87.252\% & 19457 & 0.016\% \\
8 & 16 & 32 & 88.103\% & 14977 & 0.012\% & 89.141\% & 28801 & 0.023\% \\
8 & 16 & 64 & 89.513\% & 24449 & 0.020\% & 89.820\% & 47489 & 0.038\% \\
8 & 32 & 16 & 87.523\% & 14977 & 0.012\% & 87.731\% & 28801 & 0.023\% \\
8 & 32 & 32 & 88.570\% & 19713 & 0.016\% & 88.505\% & 38145 & 0.031\% \\
8 & 32 & 64 & 89.677\% & 29185 & 0.023\% & 89.779\% & 56833 & 0.045\% \\
8 & 64 & 16 & 87.221\% & 24449 & 0.020\% & 88.249\% & 47489 & 0.038\% \\
8 & 64 & 32 & 88.112\% & 29185 & 0.023\% & 89.606\% & 56833 & 0.045\% \\
8 & 64 & 64 & 89.652\% & 38657 & 0.031\% & 89.629\% & 75521 & 0.060\% \\
16 & 32 & 32 & 89.326\% & 38657 & 0.031\% & 89.611\% & 75521 & 0.060\% \\
16 & 32 & 64 & 89.731\% & 57601 & 0.046\% & 89.761\% & 112897 & 0.090\% \\
16 & 64 & 32 & 88.807\% & 57601 & 0.046\% & 89.179\% & 112897 & 0.090\% \\
16 & 64 & 64 & 89.517\% & 76545 & 0.061\% & 89.677\% & 150273 & 0.120\% \\
32 & 64 & 64 & 89.559\% & 152321 & 0.122\% & 90.369\% & 299777 & 0.240\% \\
\hline
\end{tabular}
\caption{RoBERTa-STSB}
\end{table*}

\begin{table*}
\centering
\begin{tabular}{|c|c|c|c|c|c|c|c|c|}
\hline
\textbf{r} & \textbf{a} & \textbf{b} & \textbf{Accuracy(D)} & \textbf{Parameters(D)} & \textbf{Percentage(D)} & \textbf{Accuracy(D+A)} & \textbf{Parameters(D+A)} & \textbf{Percentage(D+A)} \\
\hline
8 & 16 & 16 & 78.619\% & 11010 & 0.010\% & 80.35\% & 20226 & 0.018\% \\
8 & 16 & 32 & 81.112\% & 15746 & 0.014\% & 79.58\% & 29570 & 0.027\% \\
8 & 16 & 64 & 80.633\% & 25218 & 0.023\% & 79.67\% & 48258 & 0.044\% \\
8 & 32 & 16 & 78.715\% & 15746 & 0.014\% & 81.21\% & 29570 & 0.027\% \\
8 & 32 & 32 & 80.633\% & 20482 & 0.019\% & 81.21\% & 38914 & 0.035\% \\
8 & 32 & 64 & 81.016\% & 29954 & 0.027\% & 79.67\% & 57602 & 0.052\% \\
8 & 64 & 16 & 77.661\% & 25218 & 0.023\% & 79.77\% & 48258 & 0.044\% \\
8 & 64 & 32 & 80.153\% & 29954 & 0.027\% & 81.59\% & 57602 & 0.052\% \\
8 & 64 & 64 & 81.783\% & 39426 & 0.036\% & 79.29\% & 76290 & 0.069\% \\
16 & 32 & 32 & 79.195\% & 39426 & 0.036\% & 82.26\% & 76290 & 0.069\% \\
16 & 32 & 64 & 81.208\% & 58370 & 0.053\% & 82.45\% & 113666 & 0.103\% \\
16 & 64 & 32 & 80.345\% & 58370 & 0.053\% & 80.54\% & 113666 & 0.103\% \\
16 & 64 & 64 & 80.825\% & 77314 & 0.070\% & 79.58\% & 151042 & 0.137\% \\
32 & 64 & 64 & 81.400\% & 153090 & 0.139\% & 80.82\% & 300546 & 0.273\% \\
\hline
\end{tabular}
\caption{BERT-CoLA}
\end{table*}

\begin{table*}
\centering
\begin{tabular}{|c|c|c|c|c|c|c|c|c|}
\hline
\textbf{r} & \textbf{a} & \textbf{b} & \textbf{Accuracy(D)} & \textbf{Parameters(D)} & \textbf{Percentage(D)} & \textbf{Accuracy(D+A)} & \textbf{Parameters(D+A)} & \textbf{Percentage(D+A)} \\
\hline
8 & 16 & 16 & 64.260\% & 11010 & 0.010\% & 64.982\% & 20226 & 0.018\% \\
8 & 16 & 32 & 58.123\% & 15746 & 0.014\% & 65.343\% & 29570 & 0.027\% \\
8 & 16 & 64 & 66.426\% & 25218 & 0.023\% & 64.621\% & 48258 & 0.044\% \\
8 & 32 & 16 & 61.733\% & 15746 & 0.014\% & 64.621\% & 29570 & 0.027\% \\
8 & 32 & 32 & 63.899\% & 20482 & 0.019\% & 67.148\% & 38914 & 0.035\% \\
8 & 32 & 64 & 61.733\% & 29954 & 0.027\% & 67.870\% & 57602 & 0.052\% \\
8 & 64 & 16 & 62.816\% & 25218 & 0.023\% & 64.260\% & 48258 & 0.044\% \\
8 & 64 & 32 & 63.538\% & 29954 & 0.027\% & 64.982\% & 57602 & 0.052\% \\
8 & 64 & 64 & 67.870\% & 39426 & 0.036\% & 67.148\% & 76290 & 0.069\% \\
16 & 32 & 32 & 62.455\% & 39426 & 0.036\% & 67.870\% & 76290 & 0.069\% \\
16 & 32 & 64 & 61.372\% & 58370 & 0.053\% & 66.065\% & 113666 & 0.103\% \\
16 & 64 & 32 & 64.260\% & 58370 & 0.053\% & 67.870\% & 113666 & 0.103\% \\
16 & 64 & 64 & 64.260\% & 77314 & 0.070\% & 64.260\% & 151042 & 0.137\% \\
32 & 64 & 64 & 61.011\% & 153090 & 0.139\% & 65.704\% & 300546 & 0.273\% \\
\hline
\end{tabular}
\caption{BERT-RTE}
\end{table*}

\begin{table*}
\centering
\begin{tabular}{|c|c|c|c|c|c|c|c|c|}
\hline
\textbf{r} & \textbf{a} & \textbf{b} & \textbf{Accuracy(D)} & \textbf{Parameters(D)} & \textbf{Percentage(D)} & \textbf{Accuracy(D+A)} & \textbf{Parameters(D+A)} & \textbf{Percentage(D+A)} \\
\hline
8 & 16 & 16 & 81.127\% & 11010 & 0.010\% & 84.559\% & 20226 & 0.018\% \\
8 & 16 & 32 & 81.373\% & 15746 & 0.014\% & 82.598\% & 29570 & 0.027\% \\
8 & 16 & 64 & 82.843\% & 25218 & 0.023\% & 81.127\% & 48258 & 0.044\% \\
8 & 32 & 16 & 79.167\% & 15746 & 0.014\% & 80.637\% & 29570 & 0.027\% \\
8 & 32 & 32 & 83.088\% & 20482 & 0.019\% & 82.353\% & 38914 & 0.035\% \\
8 & 32 & 64 & 82.598\% & 29954 & 0.027\% & 81.373\% & 57602 & 0.052\% \\
8 & 64 & 16 & 79.167\% & 25218 & 0.023\% & 84.559\% & 48258 & 0.044\% \\
8 & 64 & 32 & 81.863\% & 29954 & 0.027\% & 80.882\% & 57602 & 0.052\% \\
8 & 64 & 64 & 83.333\% & 39426 & 0.036\% & 78.922\% & 76290 & 0.069\% \\
16 & 32 & 32 & 83.088\% & 39426 & 0.036\% & 83.578\% & 76290 & 0.069\% \\
16 & 32 & 64 & 83.088\% & 58370 & 0.053\% & 79.902\% & 113666 & 0.103\% \\
16 & 64 & 32 & 84.804\% & 58370 & 0.053\% & 80.637\% & 113666 & 0.103\% \\
16 & 64 & 64 & 83.578\% & 77314 & 0.070\% & 81.373\% & 151042 & 0.137\% \\
32 & 64 & 64 & 82.598\% & 153090 & 0.139\% & 84.314\% & 300546 & 0.273\% \\
\hline
\end{tabular}
\caption{BERT-MRPC}
\end{table*}

\begin{table*}
\centering
\begin{tabular}{|c|c|c|c|c|c|c|c|c|}
\hline
\textbf{r} & \textbf{a} & \textbf{b} & \textbf{Pearson(D)} & \textbf{Parameters(D)} & \textbf{Percentage(D)} & \textbf{Pearson(D+A)} & \textbf{Parameters(D+A)} & \textbf{Percentage(D+A)} \\
\hline
8 & 16 & 16 & 85.75\% & 10241 & 0.009\% & 86.14\% & 19457 & 0.018\% \\
8 & 16 & 32 & 86.88\% & 14977 & 0.014\% & 87.53\% & 28801 & 0.026\% \\
8 & 16 & 64 & 87.75\% & 24449 & 0.022\% & 87.41\% & 47489 & 0.043\% \\
8 & 32 & 16 & 85.36\% & 14977 & 0.014\% & 86.58\% & 28801 & 0.026\% \\
8 & 32 & 32 & 86.71\% & 19713 & 0.018\% & 87.05\% & 38145 & 0.035\% \\
8 & 32 & 64 & 86.94\% & 29185 & 0.027\% & 87.86\% & 56833 & 0.052\% \\
8 & 64 & 16 & 85.86\% & 24449 & 0.022\% & 86.91\% & 47489 & 0.043\% \\
8 & 64 & 32 & 86.64\% & 29185 & 0.027\% & 87.34\% & 56833 & 0.052\% \\
8 & 64 & 64 & 86.21\% & 38657 & 0.035\% & 88.02\% & 75521 & 0.069\% \\
16 & 32 & 32 & 87.26\% & 38657 & 0.035\% & 87.64\% & 75521 & 0.069\% \\
16 & 32 & 64 & 86.63\% & 57601 & 0.052\% & 87.06\% & 112897 & 0.103\% \\
16 & 64 & 32 & 87.37\% & 57601 & 0.052\% & 87.36\% & 113666 & 0.103\% \\
16 & 64 & 64 & 88.35\% & 76545 & 0.070\% & 88.25\% & 150273 & 0.137\% \\
32 & 64 & 64 & 87.13\% & 152321 & 0.138\% & 87.98\% & 299777 & 0.273\% \\
\hline
\end{tabular}
\caption{BERT-STSB}
\end{table*}

\begin{table*}
\centering
\begin{tabular}{|c|c|c|c|c|c|c|c|c|c|c|}
\hline
\textbf{r} & \textbf{a} & \textbf{b} & \textbf{Parameters} & \textbf{Percentage} & \textbf{bleu-1} & \textbf{bleu-2} & \textbf{bleu-3} & \textbf{rouge-1} & \textbf{rouge-2} & \textbf{rouge-L} \\
\hline
8 & 16 & 16 & 20224 & 0.034\% & 31.90\% & 23.81\% & 18.17\% & 44.54\% & 26.57\% & 42.26\% \\
8 & 16 & 32 & 30336 & 0.051\% & 33.63\% & 25.39\% & 19.57\% & 46.95\% & 28.60\% & 44.69\% \\
8 & 16 & 64 & 50560 & 0.084\% & 34.23\% & 25.89\% & 20.01\% & 47.88\% & 29.42\% & 45.60\% \\
8 & 32 & 16 & 30336 & 0.051\% & 32.66\% & 24.53\% & 18.84\% & 46.01\% & 27.81\% & 43.62\% \\
8 & 32 & 32 & 40448 & 0.067\% & 34.03\% & 25.79\% & 19.94\% & 47.61\% & 29.31\% & 45.31\% \\
8 & 32 & 64 & 60672 & 0.101\% & 34.36\% & 26.13\% & 20.29\% & 48.22\% & 29.92\% & 45.88\% \\
8 & 64 & 16 & 50560 & 0.084\% & 33.19\% & 24.94\% & 19.12\% & 46.21\% & 27.83\% & 43.74\% \\
8 & 64 & 32 & 60672 & 0.101\% & 34.15\% & 25.77\% & 19.88\% & 47.80\% & 29.31\% & 45.41\% \\
8 & 64 & 64 & 80896 & 0.135\% & 34.27\% & 26.01\% & 20.15\% & 47.88\% & 29.56\% & 45.62\% \\
16 & 32 & 32 & 80896 & 0.135\% & 33.92\% & 25.60\% & 19.75\% & 47.25\% & 28.95\% & 45.00\% \\
16 & 32 & 64 & 121344 & 0.202\% & 34.54\% & 26.25\% & 20.34\% & 48.14\% & 29.77\% & 45.87\% \\
16 & 64 & 32 & 121344 & 0.202\% & 34.07\% & 25.75\% & 19.89\% & 47.66\% & 29.36\% & 45.38\% \\
16 & 64 & 64 & 161792 & 0.270\% & 34.27\% & 26.02\% & 20.19\% & 48.03\% & 29.89\% & 45.88\% \\
32 & 64 & 64 & 323584 & 0.539\% & 34.09\% & 25.82\% & 19.98\% & 47.75\% & 29.52\% & 45.54\% \\
\hline
\end{tabular}
\caption{T5-Small}
\end{table*}

\begin{table*}
\centering
\begin{tabular}{|c|c|c|c|c|c|c|c|c|c|c|}
\hline
\textbf{r} & \textbf{a} & \textbf{b} & \textbf{Parameters} & \textbf{Percentage} & \textbf{bleu-1} & \textbf{bleu-2} & \textbf{bleu-3} & \textbf{rouge-1} & \textbf{rouge-2} & \textbf{rouge-L} \\
\hline
8 & 16 & 16 & 40192 & 0.018\% & 34.63\% & 26.32\% & 20.42\% & 48.22\% & 29.76\% & 45.87\% \\
8 & 16 & 32 & 60288 & 0.027\% & 34.85\% & 26.70\% & 20.80\% & 48.79\% & 30.70\% & 46.52\% \\
8 & 16 & 64 & 100480 & 0.045\% & 34.96\% & 26.80\% & 20.89\% & 49.07\% & 30.99\% & 46.76\% \\
8 & 32 & 16 & 60288 & 0.027\% & 34.46\% & 26.11\% & 20.22\% & 48.26\% & 30.02\% & 45.93\% \\
8 & 32 & 32 & 80384 & 0.036\% & 35.04\% & 26.81\% & 20.97\% & 49.07\% & 31.08\% & 46.77\% \\
8 & 32 & 64 & 120576 & 0.054\% & 34.84\% & 26.62\% & 20.75\% & 49.10\% & 30.93\% & 46.79\% \\
8 & 64 & 16 & 100480 & 0.045\% & 34.78\% & 26.53\% & 20.63\% & 48.71\% & 30.50\% & 46.35\% \\
8 & 64 & 32 & 120576 & 0.054\% & 34.91\% & 26.73\% & 20.88\% & 48.92\% & 30.91\% & 46.69\% \\
8 & 64 & 64 & 160768 & 0.072\% & 35.16\% & 26.83\% & 20.88\% & 49.20\% & 30.94\% & 46.98\% \\
16 & 32 & 32 & 160768 & 0.072\% & 35.01\% & 26.84\% & 21.00\% & 48.95\% & 30.97\% & 46.72\% \\
16 & 32 & 64 & 241152 & 0.108\% & 34.82\% & 26.62\% & 20.77\% & 48.91\% & 30.67\% & 46.59\% \\
16 & 64 & 32 & 241152 & 0.108\% & 35.21\% & 27.05\% & 21.17\% & 49.21\% & 31.14\% & 46.98\% \\
16 & 64 & 64 & 321536 & 0.144\% & 35.18\% & 26.99\% & 21.12\% & 49.07\% & 30.98\% & 46.77\% \\
32 & 64 & 64 & 643072 & 0.288\% & 35.09\% & 26.92\% & 21.08\% & 48.96\% & 31.04\% & 46.78\% \\
\hline
\end{tabular}
\caption{T5-Base}
\end{table*}

\end{document}